\documentclass[runningheads]{llncs}

\usepackage{eccv}

\usepackage{eccvabbrv}
\usepackage{graphicx}
\usepackage{booktabs}
\usepackage[accsupp]{axessibility} 

\usepackage{algorithm}
\usepackage{algpseudocode}
\usepackage{amssymb, amsfonts}
\usepackage{amsmath}
\usepackage{tensor}
\usepackage{multirow}
\usepackage{makecell}
\usepackage{color}
\usepackage{pifont}       
\usepackage{bbding}  
\usepackage[bold]{hhtensor}
\usepackage[misc]{ifsym}

\usepackage{wrapfig}

\newcommand{\squishlist}{
 \begin{list}{$\bullet$}
  { \setlength{\itemsep}{0pt}
     \setlength{\parsep}{1pt}
     \setlength{\topsep}{1pt}
     \setlength{\partopsep}{0pt}
     \setlength{\leftmargin}{1.5em}
     \setlength{\labelwidth}{1em}
     \setlength{\labelsep}{0.5em} } }
\newcommand{\squishend}{
  \end{list}  }

\usepackage{hyperref}

\usepackage{orcidlink}

\begin{document}

\title{DeCo: Zero-Shot Industrial Anomaly Generation \\ through Decoupling and Recoupling}
\titlerunning{DeCo}

\author{Shilei Zeng \and
Xurui Li \and
Yaohan Tang \and
Yu Zhou\textsuperscript{\Letter}}

\authorrunning{S.~Zeng et al.}

\institute{School of Electronic Information and Communications,\\ 
Huazhong University of Science and Technology\\
\email{\{shlzeng,xrli\_plus,yhtang\_,yuzhou\}@hust.edu.cn}}
\newcommand{\zsl}[1]{{\color{NavyBlue} #1}}
\maketitle

\begin{abstract}
Industrial anomaly inspection is severely hindered by the scarcity of real anomalous data.
Zero-shot industrial anomaly generation addresses this by generating anomalies on specific products without requiring any of their real anomalous images. However, existing methods suffer from two critical limitations, i.e., inaccurate anomaly information acquisition
and uncontrolled anomaly-product fusion.
To overcome these challenges, we propose DeCo, 
which decouples the anomaly structure from its source product, and explicitly recouples it with the normal textures of the target product. 
During anomaly information acquisition,
Dual-Routing Flow (DR-Flow) binds the texture-invariant anomaly structure to an abnormal token, while a parallel constraint, Product-Invariant Flow (PI-Flow), prevents the abnormal token from binding the source product. 
During anomaly-product fusion, we propose a hybrid injection to recouple the acquired anomaly structure with the target product,
and Product Compatibility Correction (PCC) to compensate for the incompatibility between the acquired anomaly structure and the product. 
Extensive experiments demonstrate that DeCo establishes a new state-of-the-art. Training downstream detection models on our generated data yields massive pixel AP improvements of 5.1\% on MVTec AD and 8.2\% on VisA. Code is available at \href{https://github.com/HUST-SLOW/DeCo}{https://github.com/HUST-SLOW/DeCo}.

\end{abstract}

\section{Introduction}
\label{sec:intro}

\begin{figure}[!t]
\centering
\includegraphics[width=1\textwidth]{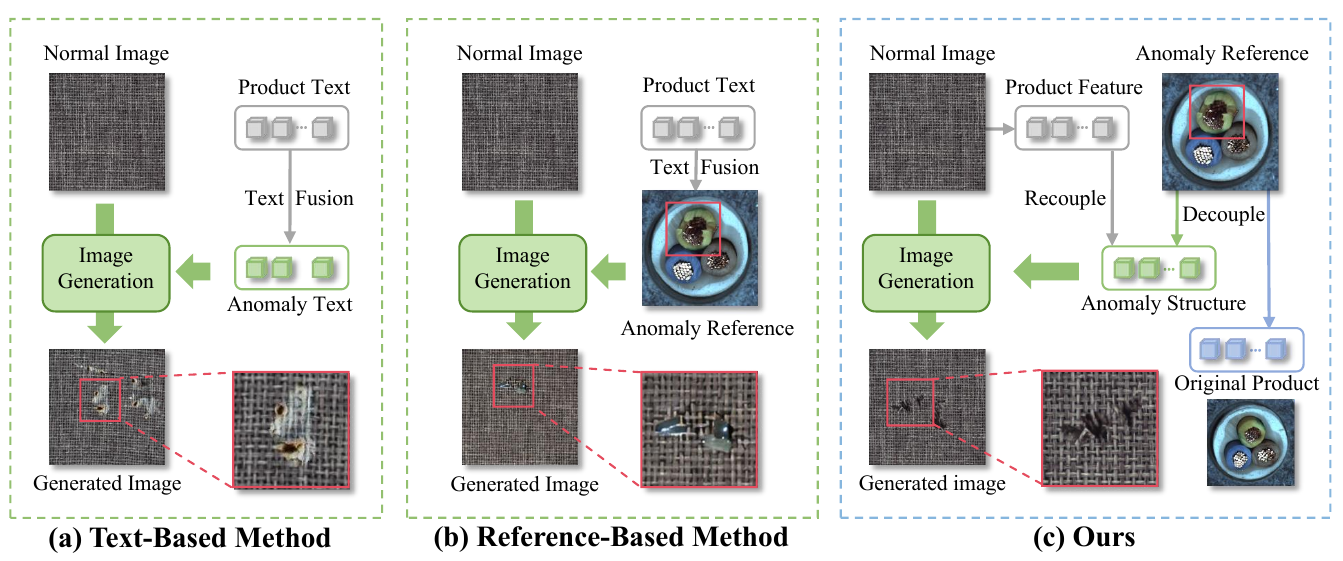}
\caption{
\textbf{(a) Text-based methods} face a semantic gap when using fixed text descriptions, yielding unrealistic patterns for complex anomalies. 
\textbf{(b) Reference-based methods} struggle to disentangle the anomaly from the original product, causing unwanted textures to leak into the target generation.
\textbf{(c)} Our method (\textbf{Ours}) explicitly decouples the anomaly structure from its original product, and seamlessly recouples it with the specific product, generating high-fidelity anomalies.
}

\label{fig:introduction}
\end{figure}

Industrial anomaly inspection is hindered by the scarcity of real-world defects.
To mitigate data scarcity, anomaly synthesis methods~\cite{devries2017cutout, li2021cutpaste, zavrtanik2021draem, schluter2022natural, zhang2024realnet} rely on data augmentation to generate pseudo-anomalies, 
but their authenticity is often limited.
Few-shot anomaly generation methods~\cite{hu2024anomalydiffusion,gui2024few,jin2024dualanodiff,dai2025seas} 
offer a promising alternative by learning from a few anomaly images. 
However, collecting even a small set of target anomalies remains challenging in cold-start scenarios.

To solve this, zero-shot industrial anomaly generation aims to generate anomalies on a specific product without requiring any real anomaly images of that product. 
Current methods primarily fall into two categories, namely, text-based~\cite{sun2025unseen,so2026anostyler} and reference-based~\cite{he2024anomalycontrol,jiang2025anomagic} generation. 
Although promising anomalies have been generated, 
they still face two main challenges,
\textbf{(1) Inaccurate Anomaly Information Acquisition.} 
Text-based methods acquire anomaly information from fixed text prompts, and use it to guide the generation. 
However, due to the semantic gap between the fixed text descriptions and the complex anomalies, these methods often generate unrealistic patterns,
as shown in Fig.~\ref{fig:introduction}(a). 
In reference-based methods, visual features are first extracted from anomalies in other products, and then used to guide generation. 
Nevertheless, they struggle to disentangle anomalies from the original product, causing unwanted textures to leak into the generated image, as shown in Fig.~\ref{fig:introduction}(b).
\textbf{(2) Uncontrolled Anomaly-Product Fusion.} 
Existing methods lack an explicit mechanism to fuse anomaly information to a specific product. They typically rely on fused text prompts (e.g., \textit{``a cut on carpet''}), which often results in poor anomaly-product fusion.

To address these challenges, we propose DeCo, a novel zero-shot industrial anomaly generation framework. As illustrated in Fig.~\ref{fig:introduction}(c), 
our core idea is to first decouple the anomaly structure from its original product context, and then explicitly recouple it with the target product.
For anomaly structure acquisition, we first introduce Dual-Routing Flow (DR-Flow), which comprises AP-Routing and QK-Routing. AP-Routing isolates the abnormal text token, 
while QK-Routing extracts the anomaly structure. 
Thus DR-Flow guarantees the anomaly structure is explicitly bound to the abnormal text token.
In addition, Product-Invariant Flow (PI-Flow) is further proposed, 
which prevents the abnormal token from directly inheriting product features from the source reference.
During anomaly-product fusion in the generation process, 
we perform hybrid injection to explicitly recouple the acquired anomaly structures with the specific product.
To address potential poor incompatibility,
we introduce Product Compatibility Correction (PCC), 
which corrects the mismatch between the anomaly structures and the product.
Extensive experiments demonstrate the superiority of DeCo in zero-shot anomaly generation. By training downstream detection models on our generated data, we establish a new state-of-the-art, outperforming existing zero-shot anomaly generation methods with a pixel AP increase of 5.1\% on the MVTec AD dataset and 8.2\% on the VisA dataset.

In summary, our contributions are threefold:
\begin{itemize}

    \item We propose DeCo, a novel zero-shot industrial anomaly generation framework. During anomaly information acquisition, it decouples the anomaly structure from its source product and explicitly binds it to an abnormal token. During anomaly-product fusion, it performs hybrid injection to explicitly recouple the acquired abnormal structure with the target product.
    
    \item The strength of DeCo stems from its multiple novel designs. DR-Flow employs AP-Routing to isolate the abnormal text token and QK-Routing to extract the anomaly structure, ensuring explicit binding between them. PI-Flow is introduced to prevent the token from inheriting source product features. At inference, Product Compatibility Correction (PCC) compensates for incompatibilities between the acquired anomaly structures and the target product.
    
    \item Extensive experiments demonstrate that DeCo establishes a new state-of-the-art for downstream zero-shot anomaly detection. Training detection models on our generated data yields significant improvements, outperforming existing methods with a pixel AP increase of 5.1\% on the MVTec AD dataset and 8.2\% on the VisA dataset.
\end{itemize}
\section{Related Work}

\label{sec:related}
\subsection{Industrial Anomaly Generation}

Industrial anomaly inspection is limited by the scarcity of real-world anomalies. 
Anomaly synthesis methods~\cite{devries2017cutout, li2021cutpaste, zavrtanik2021draem, schluter2022natural,zhang2024realnet} rely on data augmentation to generate pseudo-anomalies. 
Few-shot anomaly generation methods~\cite{Duan2023DFMGAN, hu2024anomalydiffusion, gui2024few, jin2024dualanodiff, dai2025seas} address this by generating anomalies from limited samples. 
While zero-shot anomaly generation aims to generate anomalies on a specific product without requiring any real anomaly images of that product. Existing methods face two limitations. Specifically, they struggle with anomaly information acquisition and anomaly-product fusion.
Text-driven approaches~\cite{sun2025unseen,so2026anostyler} struggle with inaccurate anomaly acquisition. 
Due to the semantic gap between the text descriptions and the anomalies, they often yield unrealistic patterns. 
Reference-based methods~\cite{he2024anomalycontrol,jiang2025anomagic} use visual features extracted from reference anomalies as guidance during generation. They also suffer from inaccurate acquisition by failing to disentangle the anomaly from its original product.
Furthermore, both text and reference-based methods exhibit uncontrolled anomaly-product fusion, lacking an explicit mechanism to seamlessly fuse the acquired anomaly information with the specific product.
In contrast, our method explicitly addresses these dual challenges. 
For anomaly information acquisition, we extract a product-invariant structure through DR-Flow and PI-Flow, leaving unwanted backgrounds behind. 
For anomaly-product fusion, we recouple the acquired anomaly structure and product via a hybrid LoRA injection, utilizing Product Compatibility Correction to address potential poor incompatibility.

\subsection{LoRA Composition in Image Generation}

LoRA composition primarily explores two directions: object integration~\cite{gu2023mix,
LoRA-Composer,jiang2024mc,dong2024continually,liu2023cones} and content-style fusion. 
The latter relies on decoupling and coupling to separate content and styles and apply them to new targets. 
Early works achieve this by merging or selectively activating pre-trained LoRAs~\cite{wu2024mixture,shah2024ziplora,ouyang2025k}.
More recently, attention has shifted to training-time decoupling. B-LoRA~\cite{frenkel2024implicit} identifies distinct roles for attention modules and achieves object-style separation. UnzipLoRA~\cite{liu2025unziplora} and QR-LoRA~\cite{yang2025qr} jointly optimize content and style LoRA, the former through parallel branch training and the latter via orthogonal constraints.
Previous LoRA-based decoupling methods rely on strong pre-trained semantics (e.g., \textit{``dog''} or \textit{``watercolor''}) to establish an initial binding for decoupling. 
In contrast, industrial anomalies are highly abstract and lack such semantic priors.
To overcome this gap, we shift from semantic-level to structural-level binding. 
Instead of relying on the text encoder's inherent knowledge, DR-Flow directly extracts the anomaly structure from the reference image and binds it to the isolated abnormal text token. 
This allows us to decouple the anomaly structure from its original product context.
\section{Preliminaries}
\label{sec:Preliminaries}
\noindent\textbf{Rectified Flow and SD3.}
Stable Diffusion 3 (SD3)~\cite{esser2024scaling} is built on Rectified Flow~\cite{liuflow}, a specific form of Flow Matching~\cite{lipmanflow}. 
Given a real image $x_0$, a VAE encoder first encodes it to a latent $z_0$.
SD3 learns a linear path $z_t = (1 - t)z_0 + t z_1$ to connect the clean latent $z_0$ and the Gaussian noise $z_1 \sim \mathcal{N}(0, I)$ across timesteps $t \in [0, 1]$.
Accordingly, the target velocity field is constant: $v_t = \frac{dz_t}{dt} = z_1 - z_0$.
SD3 utilizes a Multimodal Diffusion Transformer (MM-DiT) to predict this velocity. 
Given a text prompt $\mathcal{P}$, 
the predicted velocity $\hat{v}_\theta$ is trained using the flow matching objective:
\begin{equation}
\mathcal{L} = \mathbb{E}_{z_0, z_1, t} \left[ \| \hat{v}_\theta(z_t, t, \mathcal{P}) - (z_1 - z_0) \|_2^2 \right].
\label{eq:fm_loss}
\end{equation}
During inference, the generated latent $z_0$ is reconstructed by solving the ODE $\frac{dz_t}{dt} = \hat{v}_\theta(z_t, t, \mathcal{P})$ starting from noise $z_1$.

\noindent\textbf{LoRA Fine-tuning in SD3.}
Low-Rank Adaptation (LoRA)~\cite{LoRA} is a lightweight fine-tuning method that introduces trainable low-rank matrices into large pretrained models.  Given a frozen pretrained weight $W^0 \in \mathbb{R}^{m \times n}$, the update is parameterized by two low-rank matrices $B \in \mathbb{R}^{m \times r}$ and $A \in \mathbb{R}^{r \times n}$ with rank $r \ll \min(m, n)$:
\begin{equation}
    W = W^0 + \Delta W = W^0 + BA.
\label{eq:lora}
\end{equation}
In the MMDiT architecture of SD3, LoRA modules are typically applied to the linear projections within the attention blocks ($W^0_Q,W^0_K,W^0_V,W^0_{Out}$), then the output of each projection is augmented as:
\begin{equation}
    Y_i = (W^0_i + \Delta W_i)h, \quad i \in \{Q, K, V, Out\}, 
\end{equation}
where $h$ denotes the input hidden representation. This formulation enables efficient parameter adaptation while maintaining the knowledge of the pretrained model. For brevity, we omit the output projection $Out$ in subsequent discussions.

\noindent\textbf{Self-Attention Mechanism in Diffusion Models.}
The denoising model of SD3 is an MM-DiT~\cite{esser2024scaling}, which processes image and text tokens through a sequence of transformer blocks. Each block contains modulation mechanisms and self-attention layers.
The attention mechanism is formulated as:
\begin{equation}
    \label{eq:attention}
    \text{Attention}(Q, K, V) = \text{Softmax}\left(\frac{QK^T}{\sqrt{d}}\right) V,
\end{equation}
where $Q, K, V$ are the query, key, and value projected from the tokens.
Functionally, the Value ($V$) carries the features to be aggregated, while the Key ($K$) serves as the indexing feature for retrieval.
The Query ($Q$) determines the attention map by computing matching scores against $K$, deciding which features from $V$ are retrieved and how they are assembled.
Recent works~\cite{tumanyan2023plug, cao2023masactrl, chung2024style} have identified distinct semantic roles for query, key, and value features.
The $Q$ governs the spatial layout and structure, determining the arrangement of features,
while the $K$ and $V$ provide the appearance content to be aggregated.
\label{sec:preliminaries_attention}

\section{Method}
\label{sec:formatting}

To achieve zero-shot industrial anomaly generation, our core idea is to first decouple the reference anomaly structure from its original product context, and then explicitly recouple it with the target product.
For accurate anomaly information acquisition, 
Dual-Routing Flow (DR-Flow) is proposed to bind the anomaly structure to abnormal tokens. 
Further, we introduce Product-Invariant Flow (PI-Flow) as a parallel constraint.
It prevents the abnormal token from inheriting product features from the source reference.
During anomaly-product fusion, to recouple the acquired anomaly structure with the specific product, we perform a hybrid LoRA injection. 
We also propose Product Compatibility Correction (PCC) to address the incompatibility between the acquired anomaly and the specific product.

\subsection{Problem Statement}
We focus on zero-shot industrial anomaly generation, which aims to generate anomalies on a specific product $\mathcal{T}$ without requiring any real anomaly images of $\mathcal{T}$.
We acquire anomaly information from a reference product $\mathcal{R}$.
Formally, we have access to the reference $\mathcal{R}$, which includes a set of anomalous images $X_{\mathcal{R}}^{\text{a}} = \{x^{1}_{\text{a}}, \dots, x^{H}_{\text{a}}\}$ with corresponding masks $M_{\mathcal{R}}$, along with normal images $X_{\mathcal{R}}^{\text{n}} = \{x^{1}_{\text{n}}, \dots, x^{W}_{\text{n}}\}$.
For the specific product $\mathcal{T}$, only normal images $X_{\mathcal{T}}^{\text{n}} = \{I^{1}_{\text{n}}, \dots, I^{L}_{\text{n}}\}$ are available. No real anomaly images of $\mathcal{T}$ are used during the entire process.

To acquire the pure anomaly structure from the reference images and explicitly recouple it with the specific product, we define AP-LoRA as follows,

\noindent\textbf{A-LoRA,} stands for Anomaly LoRA. It contains a text anomaly LoRA $\Delta W^{\text{A}}_{\text{T}}$, and an image anomaly LoRA $\Delta W^{\text{A}}_{\text{I}}$, which are formally stated as follows, 
\begin{equation}
\Delta W^{\text{A}}_{\text{T}} = \{ \Delta W^{\text{A}}_{\text{TQ}}, \Delta W^{\text{A}}_{\text{TK}}, \Delta W^{\text{A}}_{\text{TV}} \}    
\end{equation}
\begin{equation}
\Delta W^{\text{A}}_{\text{I}} = \{ \Delta W^{\text{A}}_{\text{IQ}}, \Delta W^{\text{A}}_{\text{IK}}, \Delta W^{\text{A}}_{\text{IV}} \}    
\end{equation}
A-LoRA is incorporated into each block of MM-DiT following Eq. \eqref{eq:lora}, 
and is used to align abnormal tokens with anomaly structure.

\noindent\textbf{P-LoRA,} refers to the Product LoRA, comprises a text LoRA $\Delta W^{\text{P}}_\text{T}$ and an image LoRA $\Delta W^{\text{P}}_{\text{I}}$, formulated as:

\begin{equation}
\Delta W^{\text{P}}_{\text{T}} = \{ \Delta W^{\text{P}}_{\text{TQ}}, \Delta W^{\text{P}}_{\text{TK}}, \Delta W^{\text{P}}_{\text{TV}} \}    
\end{equation}
\begin{equation}
\Delta W^{\text{P}}_{\text{I}} = \{ \Delta W^{\text{P}}_{\text{IQ}}, \Delta W^{\text{P}}_{\text{IK}}, \Delta W^{\text{P}}_{\text{IV}} \}    
\end{equation}
Similarly, we integrate P-LoRA into the MM-DiT blocks, and establish the alignment between normal products and their text descriptions.

\begin{figure}[!t]
\centering
\includegraphics[width=1\textwidth]{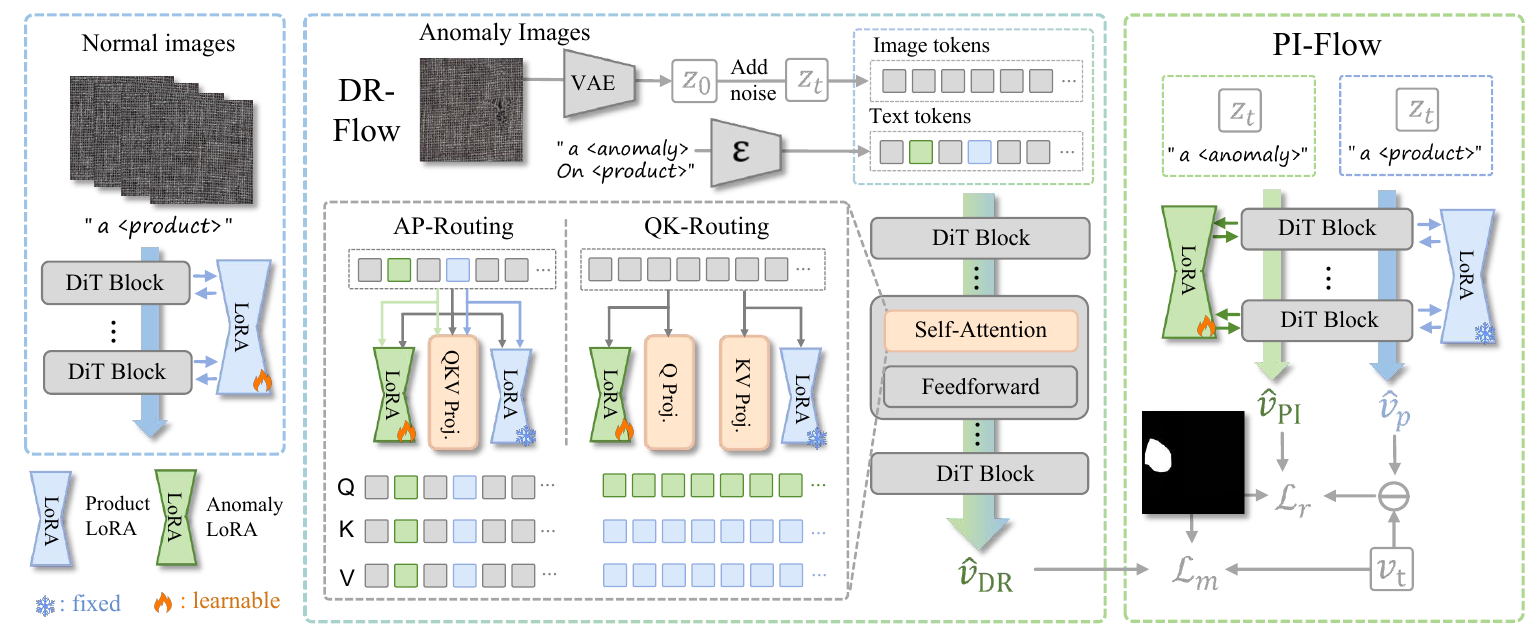}
\caption{\textbf{Training pipeline of DeCo.}
It employs a Dual-LoRA design with a frozen Product LoRA (blue, pre-trained on the \textbf{left}) and a trainable Anomaly LoRA (green).
\textbf{DR-Flow (Middle)} binds the anomaly structure to the abnormal token via AP-Routing and QK-Routing.
\textbf{PI-Flow (Right)} acts as a constraint to prevent the abnormal token from directly inheriting product features from the source reference.
}
\label{fig:gen}
\end{figure}

\subsection{Anomaly Structure Binding via Dual-Routing Flow}
\label{sec:AG-Flow}
The key to accurate anomaly acquisition lies in decoupling the anomaly structure from the reference images and binding it to the text tokens.
To this end, we introduce two mechanisms to train A-LoRA: 
an AP-Routing that separates text tokens to isolate the abnormal token, and a QK-Routing that restricts A-LoRA to mapping only the structure features.
Thus DR-Flow effectively binds the extracted anomaly structure to the isolated abnormal token.

\noindent\textbf{Product Information Binding.}
As shown in Fig.\ref{fig:gen}, given the normal product text prompt (e.g., \textit{``a <product>''}) and the normal image set $X_{\mathcal{R}}^{\text{n}}$, we train P-LoRA ($\Delta W^{\text{P}}_{\text{I}}$ and $\Delta W^{\text{P}}_{\text{T}}$) following Eq. \eqref{eq:fm_loss} to bind $X_{\mathcal{R}}^{\text{n}}$ to normal product text prompt. 
This step explicitly binds the normal product features to product text via P-LoRA,
and we freeze P-LoRA in all subsequent processing.

\noindent\textbf{AP-Routing for Text Token Split.}
Given a text prompt $\mathcal{P}$ such as \textit{``a <anomaly> on <product>''}, 
SD3 first tokenizes $\mathcal{P}$ to form the text tokens $\text{T} \in \mathbb{R}^{N \times C}$, where $N$ is the number of text tokens and $C$ is the channel number.
Then, we decompose $\text{T}$ into three groups, 
namely, $\text{T}_\text{A} \in \mathbb{R}^{1 \times C}$, $\text{T}_\text{P} \in \mathbb{R}^{1 \times C}$ and $\text{T}_\text{O} \in \mathbb{R}^{(N-2) \times C}$, which indicate the abnormal text \textit{``<anomaly>''}, 
product text \textit{``<product>''} and other texts like \textit{``a''}, \textit{``on''} and padding tokens.
Accordingly, the corresponding $Q_{\text{T}}$, $K_{\text{T}}$, and $V_{\text{T}}$ are computed using the weights $W_{\text{T}} = \{W^{\text{A}}_{\text{T}}, W^{\text{P}}_{\text{T}}, W^{\text{O}}_{\text{T}} \}$ as follows:

\begin{equation}
\begin{gathered}
    (Q^{\text{A}}_{\text{T}}, K^{\text{A}}_{\text{T}}, V^{\text{A}}_{\text{T}}) = W^{\text{A}}_{\text{T}} \, \text{T}_{\text{A}} = (W^{0}_{\text{T}} + \Delta W^{\text{A}}_{\text{T}}) \, \text{T}_{\text{A}}, \\
    (Q^{\text{P}}_{\text{T}}, K^{\text{P}}_{\text{T}}, V^{\text{P}}_{\text{T}}) = W^{\text{P}}_{\text{T}} \, \text{T}_{\text{P}} = (W^{0}_{\text{T}} + \Delta W^{\text{P}}_{\text{T}}) \, \text{T}_{\text{P}}, \\
    (Q^{\text{O}}_{\text{T}}, K^{\text{O}}_{\text{T}}, V^{\text{O}}_{\text{T}}) = W^{\text{O}}_{\text{T}} \, \text{T}_{\text{O}} = (W^{0}_{\text{T}} + \Delta W^{\text{A}}_{\text{T}} + \Delta W^{\text{P}}_{\text{T}}) \, \text{T}_{\text{O}}.
\end{gathered}
\label{eq:text_kqv}
\end{equation}
where $W^{0}_{\text{T}}$ is the pre-trained weight, and text P-LoRA $\Delta W^{\text{P}}_{\text{T}}$ is learned on product $\mathcal{R}$. Both of them are fixed in \cref{eq:text_kqv}, and the text A-LoRA $\Delta W^{\text{A}}_{\text{T}}$ is learnable. 
We then concatenate $Q^{\text{A}}_{\text{T}} \in \mathbb{R}^{1 \times C}$, $Q^{\text{P}}_{\text{T}}\in \mathbb{R}^{1 \times C}$, and $Q^{\text{O}}_{\text{T}} \in \mathbb{R}^{(N-2) \times C}$, 
resulting in $Q_{\text{T}} \in \mathbb{R}^{N \times C}$.
We obtain $K_{\text{T}} \in \mathbb{R}^{N \times C}$ and $V_{\text{T}} \in \mathbb{R}^{N \times C}$ in the same way.
AP-Routing maps the anomaly and product tokens separately, ensuring that each of them is processed exclusively by its own LoRA. 
We explicitly separate them because industrial anomaly structure carries limited semantic information, unlike the objects in natural scenes (e.g., \textit{``dog''}). 
By isolating the text routing, we ensure that the abstract anomaly structure is bound solely to the abnormal text token, independent of the product context.

\noindent\textbf{QK-Routing for Structure Extraction.}
Given the image token $\text{T}_{\text{I}} \in \mathbb{R}^{M \times C}$,
we expand the pre-trained weight $W^{0}_{\text{I}}$ as $\{W^{0}_{\text{IQ}}, W^{0}_{\text{IK}}, W^{0}_{\text{IV}}\}$,   
and propose a QK-Routing mechanism as follows,
\begin{equation}
\begin{gathered}
    Q_{\text{I}} = (W^{0}_{\text{IQ}} + \Delta W^{\text{A}}_{\text{IQ}}) \text{T}_{\text{I}}, \\
    K_{\text{I}} = (W^{0}_{\text{IK}} + \Delta W^{\text{P}}_{\text{IK}}) \text{T}_{\text{I}}, \\
    V_{\text{I}} = (W^{0}_{\text{IV}} + \Delta W^{\text{P}}_{\text{IV}}) \text{T}_{\text{I}}
\end{gathered}
\label{eq:img_kqv}
\end{equation}
In \cref{eq:img_kqv}, we map $\text{T}_{\text{I}}$ to $K$ and $V$ using the frozen P-LoRA parameters.
As discussed in Sec.~\ref{sec:preliminaries_attention}, the query ($Q$) controls the spatial structure, while the key ($K$) and value ($V$) provide the visual texture. 
By computing $K_{\text{I}}$ and $V_{\text{I}}$ using only the frozen P-LoRA, we limit the available textures to the normal product. 
At the same time, we compute $Q_{\text{I}}$ using the trainable A-LoRA. 
Through this design, the trainable A-LoRA extracts the reference anomaly's structure into $Q_{\text{I}}$.

Within the standard MM-DiT block, the queries ($Q_{\text{T}}$ and $Q_{\text{I}}$), keys ($K_{\text{T}}$ and $K_\text{I}$), and values ($V_{\text{T}}$ and $V_\text{I}$) are concatenated to construct a unified set $\{Q_\text{T,I}, K_\text{T,I}, V_\text{T,I}\}$ for joint self-attention.
During this process, the isolated abnormal text token acts as the sole text condition for the A-LoRA. 
Driven by the generation loss, this attention interaction inherently forces A-LoRA to bind the anomaly structure to the abnormal text token.
After the joint self-attention, these features are processed by the remaining MM-DiT blocks to predict the velocity $\hat{v}_{\text{DR}}$.
The output $\hat{v}_{\text{DR}}$ is supervised using a velocity-based flow-matching loss restricted to the anomaly region:
\begin{equation}
  \mathcal{L}_{\text{DR}} = \mathbb{E}_{z_0, z_1, t, m} \left[ \left\| m \odot (\hat{v}_{\text{DR}}(z_t, t, \mathcal{P}) - v_t) \right\|_2^2 \right],
\end{equation}
where $\hat{v}_{\text{DR}}$ is the predicted velocity and $v_t = z_1 - z_0$ is the constant target velocity. The variable $z_t$ represents the noise latent at timestep $t$, $m$ is the binary anomaly mask, and $\mathcal{P}$ denotes the condition prompt.

\subsection{Product-Invariant Flow}
\label{sec:RD-Reg}

\begin{figure}[!t]
\centering
\includegraphics[width=1\textwidth]{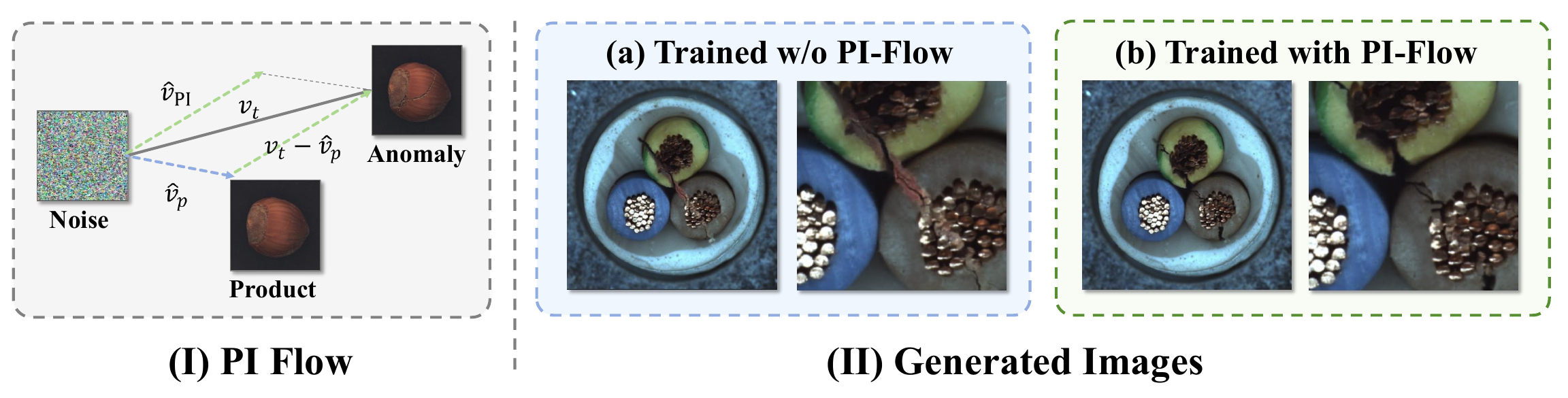}
\caption{
(I) PI-Flow. (II) Without PI-Flow, the generated anomalies may carry over the source product’s appearance.
}
\label{fig:PI-Flow}
\vspace{-5pt}
\end{figure}

In DR-Flow, we train A-LoRA to bind the anomaly structure to the abnormal token. However, to reduce the training loss, A-LoRA might simply align the abnormal token to the original reference anomaly that contains the product's texture. 
As a result, the generated anomaly carries over the source product's appearance, as shown in Fig.~\ref{fig:PI-Flow}(II)(a).

To further ensure accurate and pure anomaly structure acquisition, we introduce Product-Invariant Flow (PI-Flow) as a parallel constraint.
This constraint prevents the abnormal token from binding the source product. 
Specifically, given the prompt \textit{``a <product>''}, and the noise anomaly latent $z_t$ as input,
we activate the fixed P-LoRA alone
to predict a product flow $\hat{v}_p$ in the anomaly region, thus PI-Flow is stated as follows:
\begin{equation}
    v_{\text{PI}} = v_t - \hat{v}_p
\end{equation}
As shown in Fig.~\ref{fig:PI-Flow}(I), since P-LoRA is fixed and trained on normal images, $\hat{v}_p$ is a constant velocity that solely reconstructs the normal product. 
By explicitly subtracting this product-specific velocity $\hat{v}_p$ from the total target velocity $v_t$, $v_{\text{PI}}$ acts as a residual flow. 
This residual flow represents the pure anomaly, completely separated from the normal product.

In addition, given the prompt $\mathcal{P}_a$ (\textit{``a <anomaly>''}) and the noise anomaly latent $z_t$ as input, we activate the trainable A-LoRA $\Delta W^{\text{A}}$ alone to predict the product-invariant flow $\hat{v}_{\text{PI}}$. The objective is defined as follows: 
\begin{equation}
    \mathcal{L}_{\text{PI}} = \mathbb{E}_{z_0, z_1, t, m} \left[ \left\| m \odot \left( \hat{v}_{\text{PI}}(z_t,t,\mathcal{P}_a) - v_{\text{PI}} \right) \right\|_2^2 \right].
\end{equation}
Consequently, $\Delta W^{\text{A}}$ is explicitly constrained to bind these product-invariant anomaly features to the abnormal token. This prevents the token from inheriting product features, as illustrated in Fig.~\ref{fig:PI-Flow}(II)(b).

The total training objective is defined as:
\begin{equation}
    \mathcal{L}_{\text{total}} = \mathcal{L}_{\text{DR}} + \lambda_{\text{PI}} \cdot \mathcal{L}_{\text{PI}}.
\end{equation}
Here, $\mathcal{L}_{\text{DR}}$ drives the text-structure binding within the DR-Flow.
Furthermore, $\mathcal{L}_{\text{PI}}$ prevents the abnormal tokens from binding the source product. By adjusting the weighting factor $\lambda_{\text{PI}}$, we control the strength of this constraint. 
Guided by this joint objective, the A-LoRA is forced to bind product-invariant anomaly structure to the abnormal token.

\subsection{Inference with Compatibility Correction}
\label{sec:Inference}

During the generation stage, 
given the pretrained A-LoRA $\Delta W_{\mathcal{R}}^{\text{A}}$ on reference product $\mathcal{R}$,
we aim to recouple the acquired anomaly structure from A-LoRA with the specific product $\mathcal{T}$.
To this end, we adopt the following hybrid injection strategy.

\noindent\textbf{SD3 with hybrid LoRA injection.} 
We first build the inference model upon the frozen SD3 backbone weight $W_0$. 
To recouple the acquired anomaly structure with the specific product $\mathcal{T}$, we inject both the P-LoRA $\Delta W_{\mathcal{T}}^{\text{P}}$ of the specific product $\mathcal{T}$ and the A-LoRA $\Delta W_{\mathcal{R}}^{\text{A}}$ from the source product $\mathcal{R}$ and form the model weight $W_{\text{inf}}$ as follow:
\begin{equation}
    W_{\text{inf}} = W_0 + \Delta W_{\mathcal{T}}^{\text{P}} + \Delta W_{\mathcal{R}}^{\text{A}}.
\label{eq:inject}
\end{equation}
We apply Eq.\ref{eq:inject} to each block of MM-DiT, resulting in the final inference model.
 
\noindent\textbf{Hybrid Text Prompt.}
To guide the generation, we use a composite prompt based on the anomaly type from $\mathcal{R}$ and the product category of $\mathcal{T}$.
Given the text prompt template \textit{``a <anomaly> on <product>''}, we combine the name of acquired anomaly, e.g., \textit{``cut''}, and the product name of $\mathcal{T}$, e.g., \textit{``cable''}, 
resulting in a new composite prompt $\mathcal{P}_\text{mix}$, e.g., \textit{``a cut on cable''}. 

\begin{figure}[!t]
\centering
\includegraphics[width=1\textwidth]{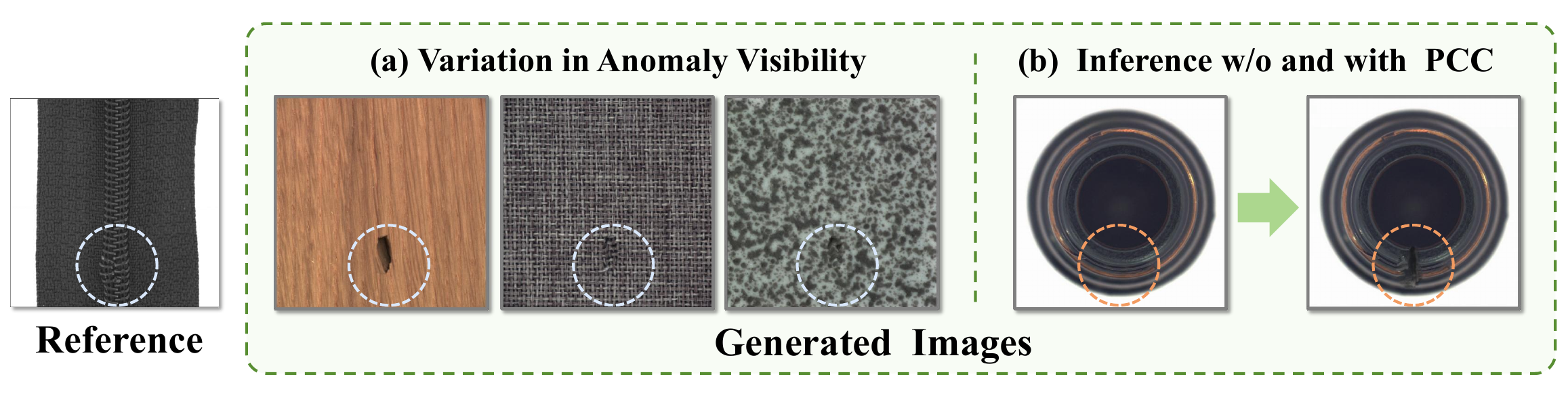}
\caption{
Zero-shot anomaly generation using the same reference anomaly. (a) Across different products, the anomaly varies in visibility. (b) With PCC to address the incompatibility, the anomaly can be generated well.
}
\label{fig:PCC}
\end{figure}

\noindent\textbf{Product Compatibility Correction.}
While the hybrid injection enables zero-shot anomaly generation, recoupling the acquired anomaly structure with different specific products $\mathcal{T}$ often reveals varying degrees of compatibility.
This variation arises because the two LoRAs are trained independently on different products.
As a result, the same acquired anomaly can exhibit different levels of effectiveness when recoupled with different products.  
As shown in Fig.~\ref{fig:PCC}(a),
in some cases, the anomaly is injected with varying visibility. 
For certain products, however,
the injected anomaly structure exhibits incompatibility with the product, making the anomaly barely visible.
This creates a misleading impression that the anomaly cannot be generated.

To address this incompatibility,
we propose Product Compatibility Correction (PCC). This strategy refines the velocity prediction.
Specifically, we compute a product-only velocity $\hat{v}_p$ by activating $\Delta W_{\mathcal{T}}^{\text{P}}$ alone. We then push the inference model prediction $\hat{v}_{\text{DR}}$ away from the product-only prediction to obtain the final inference velocity:
\begin{equation}
    \hat{v}_{\text{inf}} = \hat{v}_p + \omega \cdot (\hat{v}_{\text{DR}} - \hat{v}_p).
\end{equation}
Here, $\omega$ is a scalar hyperparameter that controls the degree of correction.
This extrapolation resembles Classifier-Free Guidance (CFG)~\cite{ho2021classifier}, which amplifies the difference between conditional and unconditional predictions to improve prompt fidelity. 
In our case, $\hat{v}_p$ represents the normal product. 
By pushing the final prediction away from this product-only prediction,
we effectively compensate for the incompatibility between the acquired anomaly structure and the product, 
enabling the successful generation of anomalies on the specific product $\mathcal{T}$, as shown in Fig.\ref{fig:PCC}(b).

Therefore, given a normal image $I_{\text{n}}$ of $\mathcal{T}$, an anomaly mask $m$, and the composite prompt $\mathcal{P}_\text{mix}$,
we first encode $I_{\text{n}}$ into latent $z_{\text{n}}$ through VAE encoder. Then we use the composite prompt $\mathcal{P}_\text{mix}$ as guidance and inject anomalies via a blended diffusion~\cite{avrahami2022blended}:
\begin{equation}
    z_t = \hat{z}_t \odot m + z_t^{\text{ref}} \odot (1 - m),
\end{equation}
where $\hat{z}_t$ is the denoised latent obtained using $\hat{v}_{\text{inf}}$, and $z_t^{\text{ref}}$ is the corresponding noisy latent obtained by forward noising $z_{\text{n}}$ to timestep $t$. This blending ensures that the acquired anomaly is injected only within the masked region, while preserving the original product elsewhere.

\section{Experiments}
\label{sec:experiments}

\subsection{Experimental setting}

\noindent\textbf{Datasets.}
We conduct our experiments on the MVTec AD dataset \cite{bergmann2019mvtec}. 
The MVTec AD dataset comprises 15 product categories, including 5 texture types and 10 object types, with each category containing up to 8 different types of anomalies,
making it suitable for simulating real-world industrial scenarios. For a more extensive evaluation, we also include the VisA dataset \cite{zou2022visa}, which consists of 12 objects spanning 3 distinct domains. 

\noindent\textbf{Implementation Details.}
We build our framework upon the pre-trained Stable Diffusion 3 (SD3). 
In our zero-shot setting, for the specific product $\mathcal{T}$, no real anomaly images are available. 
To generate anomalies for $\mathcal{T}$, we use the other 14 products of MVTec AD as our reference set. 
During the anomaly information acquisition phase, we extract product-invariant anomaly structures by training a dedicated A-LoRA for each of these 14 reference products, setting the PI-Flow loss weight $\lambda_{\text{PI}} = 0.1$. 
During generation, we explicitly recouple the normal images of $\mathcal{T}$ with the anomaly structures acquired from this reference set. 
For each category, we generate 1,000 images.
To strictly preserve the zero-shot paradigm, we uniformly utilize reference masks without any similarity-based filtering. 
The conditional anomaly masks are derived from the reference ground-truth masks and augmented through mask combinations, specifically by taking the union of two randomly sampled masks of the same anomaly type. Following~\cite{gui2024few}, these combined masks are intersected with the specific product's foreground mask to ensure valid anomaly placement. The PCC scale $\omega$ is set to 2. More details are in the Appendix \ref{app:a}.

\noindent\textbf{Baselines.}
We evaluate our method against a diverse set of anomaly synthesis and generation methods. 
For visual generation quality, 
we compare our method with anomaly synthesis methods, including CutPaste~\cite{li2021cutpaste}, DRAEM~\cite{zavrtanik2021draem}, NSA~\cite{schluter2022natural}, RealNet~\cite{zhang2024realnet},
and zero-shot anomaly generation methods AnomalyAny~\cite{sun2025unseen} and AnoStyler~\cite{so2026anostyler}. 
We also compare our method with some few-shot anomaly generation methods~\cite{duan2023defect, hu2024anomalydiffusion, jin2024dualanodiff}.
For downstream anomaly detection tasks, we compare our approach with anomaly synthesis and zero-shot anomaly generation methods. 
Additionally, to validate our advantages, we compare our approach with recent state-of-the-art Dual-LoRA decoupling and coupling methods (UnzipLoRA~\cite{liu2025unziplora} and QR-LoRA~\cite{yang2025qr}).

\noindent\textbf{Evaluation Metrics.}
For image generation, following~\cite{hu2024anomalydiffusion}, we use Inception Score (IS)~\cite{salimans2016improved} and Intra-cluster pairwise LPIPS distance (IC-LPIPS)~\cite{ojha2021fewshot} to evaluate the authenticity and diversity of the anomaly images. 
Furthermore, to specifically assess the local authenticity of the generated anomalies, we compute the IS exclusively within the anomaly regions, denoted as IS(a).
For the downstream anomaly detection task, we use Area Under ROC Curve (AUROC), Average Precision (AP), and F1-score at both the image level (I-AUC, I-AP, I-F1) and pixel level (P-AUC, P-AP, P-F1). 
Additionally, we include Per-Region Overlap (PRO) to evaluate pixel-level detection performance.

\begin{table*}[t]
\centering
\caption{Quantitative comparison of anomaly image generation quality on the MVTec AD dataset. We evaluate authenticity via IS and IS(a), and diversity via IC-L.}
\tabcolsep=3pt 
\renewcommand{\arraystretch}{1} 

\resizebox{\linewidth}{!}{%
\begin{tabular}{l|ccc|cccc|ccc}
\toprule\toprule
\multirow{2}{*}{\textbf{Metric}} & 
\multicolumn{3}{c|}{\textit{Few-Shot Generation}} & 
\multicolumn{4}{c|}{\textit{Anomaly Synthesis}} & 
\multicolumn{3}{c}{\textit{Zero-Shot Generation}} \\
\cmidrule(lr){2-4} \cmidrule(lr){5-8} \cmidrule(lr){9-11}
 & DFMGAN & AnoDiff & AnoGen & CutPaste & DRAEM & NSA & RealNet & AnomalyAny & AnoStyler & \textbf{Ours}\\
\midrule
IS $\uparrow$    & 1.72 & 1.80 & 1.77  & 1.76 & 1.76 & 1.44 & 1.64 & 2.02 & 2.04 & \textbf{2.06} \\
IS(a) $\uparrow$ & 2.80 & 2.77 & 2.84  & -- & -- & -- & -- & 2.98 & 3.12 & \textbf{3.98} \\
IC-L $\uparrow$  & 0.20 & 0.27 & 0.32  & 0.22 & 0.25 & 0.26 & 0.22 & \textbf{0.33} & 0.32 & \textbf{0.33} \\
\bottomrule\bottomrule
\end{tabular}}
\label{tab:generation}
\end{table*}

\begin{table}[t]
\centering
\caption{Comparison of anomaly detection with zero-shot anomaly synthesis and generation methods and decoupling methods on MVTec AD.}
\tabcolsep=11pt
\renewcommand{\arraystretch}{1}
\resizebox{\linewidth}{!}{%
\begin{tabular}{l|ccc|cccc}
\toprule\toprule
\multirow{2}{*}{\textbf{\large Method}} & \multicolumn{3}{c|}{\textbf{Image-Level}} & \multicolumn{4}{c}{\textbf{Pixel-Level}} \\
\cmidrule(lr){2-4} \cmidrule(lr){5-8}
 & \textbf{I-AUC} & \textbf{I-AP} & \textbf{I-F1} & \textbf{P-AUC} & \textbf{P-AP} & \textbf{P-F1} & \textbf{PRO} \\
\midrule
\multicolumn{8}{l}{\textit{Zero-shot Anomaly Synthesis and Generation Methods}} \\
\midrule
CutPaste~\cite{li2021cutpaste}     & 89.8 & 92.1 & 89.8 & 88.2 & 51.9 & 50.7 & 76.4 \\
DRAEM~\cite{zavrtanik2021draem}    & 94.6 & 97.0 & 94.4 & 92.2 & 54.1 & 53.1 & 83.1 \\
NSA~\cite{schluter2022natural}     & 93.0 & 95.6 & 91.6 & 92.0 & 52.6 & 52.5 & 82.2 \\
RealNet~\cite{zhang2024realnet}    & 95.2 & 97.0 & 95.3 & 94.0 & 57.7 & 56.6 & 85.2 \\
AnomalyAny~\cite{sun2025unseen}    & 95.2 & 96.9 & 96.3 & 89.0 & 62.7 & 59.9 & 84.7 \\
AnoStyler~\cite{so2026anostyler}   & 98.0 & 99.0 & 97.0 & 94.4 & 62.9 & 60.7 & 88.3 \\
\midrule
\multicolumn{8}{l}{\textit{Dual-LoRA Decoupling Methods}} \\
\midrule
UnzipLoRA~\cite{liu2025unziplora}  & 96.6 & 98.5 & 96.6 & 94.6 & 61.9 & 59.2 & 85.0 \\
QR-LoRA~\cite{yang2025qr}          & 95.4 & 97.2 & 95.3 & 93.1 & 60.2 & 58.3 & 84.2 \\
\midrule
\textbf{Ours}               & \textbf{98.0} & \textbf{99.1} & \textbf{96.7} & \textbf{95.3} & \textbf{68.0} & \textbf{64.5} & \textbf{89.6} \\
\bottomrule\bottomrule
\end{tabular}
}
\label{tab:mvtec_merged}
\end{table}

\subsection{Comparison in Anomaly Image Generation}
\begin{figure}[!t]
\centering
\includegraphics[width=1\textwidth]{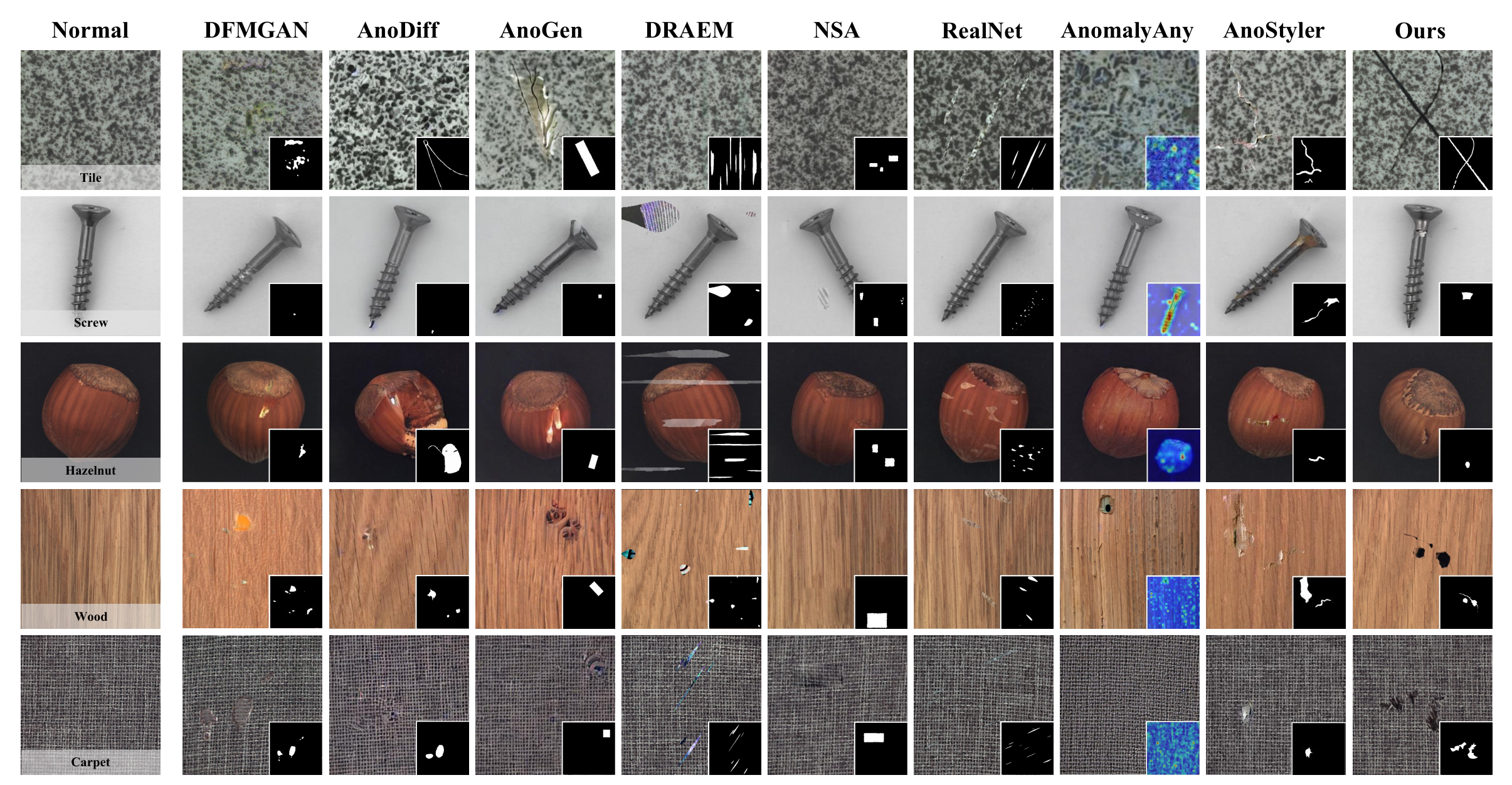}
\caption{
Comparison of generated anomaly images.
}
\label{fig:gen_result}
\end{figure}

We evaluate the authenticity and diversity of the generated anomalies in Tab.~\ref{tab:generation}. Our method achieves the highest IS (2.06) and IC-L (0.33). Additionally, for the evaluation of anomaly authenticity, our approach obtains an IS(a) of 3.98. This indicates that the anomalies generated by our method possess higher authenticity. With the decoupled structure, our method generates diverse anomalies while keeping them realistic, achieving better results than several few-shot methods.
Fig.~\ref{fig:gen_result} shows the visual comparisons. Anomaly synthesis methods (e.g., DRAEM, NSA) produce obvious artifacts, such as unnatural stripes on the hazelnut. Few-shot methods (e.g., DFMGAN, AnoGen) lack realism and show unnatural blending. Recent zero-shot methods like AnomalyAny and AnoStyler struggle with faithful anomalies (e.g., tile and wood). In contrast, our method generates highly authentic anomalies. The generated anomaly seamlessly blends with the product textures (e.g., screw, carpet). More generated images are provided in Appendix \ref{app:d}.

\subsection{Comparison in Downstream Anomaly Detection Tasks}

Following~\cite{hu2024anomalydiffusion}, we train a U-Net~\cite{ronneberger2015u} segmentation model on the generated image-mask pairs to evaluate downstream performance on the MVTec AD dataset, with results summarized in Tab.~\ref{tab:mvtec_merged}. By establishing text-structure binding via DR-Flow, our method achieves 68.0\% in P-AP and 64.5\% in P-F1, outperforming the zero-shot method AnoStyler by 5.3\% and 3.8\%. It also obtains a PRO score of 89.6\%. Furthermore, by explicitly binding the anomaly structure to the abnormal token, our method yields a 6.1\% P-AP improvement over the Dual-LoRA decoupling method. For image-level metrics, our method also achieves the highest performance, with 98.0\% I-AUC and 99.1\% I-AP. 
These results indicate that the generated zero-shot anomalies are realistic and effectively contribute to downstream detection tasks.

\subsection{Generalization of Acquired Anomaly Structures}
To further verify the generalization of the anomaly structures acquired by our method, we introduce 12 distinct and highly complex product categories of the VisA dataset~\cite{zou2022visa}. Then, we directly recouple the anomaly structures previously acquired from the MVTec AD dataset with these products. This design tests whether the structural priors acquired from the reference set can be seamlessly recoupled with the normal textures of distinct product categories. As shown in Tab.~\ref{tab:visa_detection}, our method consistently outperforms existing methods. While maintaining state-of-the-art image-level metrics (95.9\% I-AUC), our structural decoupling and explicit recoupling strategy yields massive gains in pixel-level localization. Most notably, our method reaches 41.5\% in P-AP and 46.4\% in P-F1, eclipsing the strongest method by absolute margins of 8.2\% and 5.4\%, respectively. This confirms that our framework successfully extracts generalizable anomaly structures, enabling robust zero-shot generation. More results on other datasets are shown in Appendix \ref{app:c}.

\begin{table}[!t]
\centering
\caption{Comparison of anomaly detection on VisA.}
\tabcolsep=11pt 
\renewcommand{\arraystretch}{1}
\resizebox{\linewidth}{!}{%
\begin{tabular}{l|ccc|cccc}
\toprule\toprule
\multirow{2}{*}{\textbf{\large Method}} & \multicolumn{3}{c|}{\textbf{Image-Level}} & \multicolumn{4}{c}{\textbf{Pixel-Level}} \\
\cmidrule(lr){2-4} \cmidrule(lr){5-8}
 & \textbf{I-AUC} & \textbf{I-AP} & \textbf{I-F1} & \textbf{P-AUC} & \textbf{P-AP} & \textbf{P-F1} & \textbf{PRO} \\
\midrule
CutPaste~\cite{li2021cutpaste}     & 86.3 & 86.9 & 87.1 & 88.4 & 32.2 & 39.6 & 77.7 \\
DRAEM~\cite{zavrtanik2021draem}    & 91.8 & 92.9 & 88.6 & 91.4 & 29.5 & 37.2 & 81.9 \\
NSA~\cite{schluter2022natural}     & 87.3 & 89.8 & 84.2 & 92.6 & 26.1 & 34.2 & 74.2 \\
RealNet~\cite{zhang2024realnet}    & 92.6 & 93.8 & 89.2 & 92.2 & 33.3 & 41.0 & 83.0 \\
AnomalyAny~\cite{sun2025unseen}    & 88.9 & 86.2 & 85.9 & 90.4 & 31.2 & 33.0 & 84.6 \\
AnoStyler~\cite{so2026anostyler}   & 93.9 & 95.3 & 90.1 & 93.8 & 31.4 & 36.4 & 84.3 \\
\midrule
\textbf{Ours}               & \textbf{95.9} & \textbf{97.0} & \textbf{92.1} & \textbf{96.0} & \textbf{41.5} & \textbf{46.4} & \textbf{88.3} \\
\bottomrule\bottomrule
\end{tabular}%
}
\label{tab:visa_detection}
\end{table}

\subsection{Ablation Studies}
Tab.~\ref{tab:ablation_main} presents the impact of each core component on MVTec AD. Starting from a unified LoRA baseline (a), we incrementally add AP-Routing (short for AP), QK-Routing (short for QK), PI-Flow, and PCC. Introducing AP-Routing (b) significantly improves image-level detection by ensuring token split. Adding QK-Routing (c) enhances all metrics by enabling structure extraction. Incorporating PI-Flow (d) further improves image-level and pixel-level precision by preventing the abnormal token from binding the source product. Finally, the full method (e) yields the best overall performance. These results validate the effectiveness and complementarity of each proposed component. 
More ablation studies are provided in Appendix \ref{app:b}.

\begin{table}[!t]
\centering
\caption{Ablation study of core components.}
\label{tab:ablation_main}
\tabcolsep=7pt 
\renewcommand{\arraystretch}{1.3}

\resizebox{\linewidth}{!}{%
\begin{tabular}{c|cccc|cccccc}
\toprule\toprule
\multicolumn{1}{c|}{\multirow{2}{*}{\textbf{Row}}} & \multicolumn{4}{c|}{\textbf{Components}} & \multicolumn{3}{c}{\textbf{Image-Level}} & \multicolumn{3}{c}{\textbf{Pixel-Level}} \\
\cmidrule(lr){2-5} \cmidrule(lr){6-8} \cmidrule(lr){9-11}
 & \textbf{AP} & \textbf{QK} & \textbf{PI-Flow} & \textbf{PCC} & \textbf{I-AUC} & \textbf{I-AP} & \textbf{I-F1} & \textbf{P-AUC} & \textbf{P-AP} & \textbf{P-F1} \\
\midrule
(a) & - & - & - & - & 94.0 & 97.3 & 95.4 & 94.2 & 62.7 & 60.5 \\
(b) & \checkmark & - & - & - & 95.7 & 98.0 & 96.2 & 94.3 & 63.2 & 61.1 \\
(c) & \checkmark & \checkmark & - & - & 96.9 & 98.8 & 96.3 & 95.0 & 66.7 & 63.0 \\
(d) & \checkmark & \checkmark & \checkmark & - & 97.5 & 99.0 & 96.4 & 95.1 & 67.6 & 63.2 \\
(e) & \checkmark & \checkmark & \checkmark & \checkmark & \textbf{98.0} & \textbf{99.1} & \textbf{96.7} & \textbf{95.3} &\textbf{ 68.0} & \textbf{64.5} \\
\bottomrule\bottomrule
\end{tabular}%
}
\end{table}

\section{Conclusion}
\label{sec:conclusion}
In this paper, we propose DeCo to address the critical challenges of zero-shot industrial anomaly generation. 
DeCo explicitly decouples the anomaly structure from its source product and recouples it with the target product. Specifically, DR-Flow and PI-Flow work collaboratively to extract a pure anomaly structure. 
Then, our hybrid injection and Product Compatibility Correction seamlessly fuse this acquired anomaly structure into the target product. Extensive experiments confirm that DeCo establishes a new state-of-the-art. Training downstream detection models on our generated data yields pixel AP improvements of 5.1\% on MVTec AD and 8.2\% on VisA.

\section{Acknowledgement}
This work was supported by the National Natural Science Foundation of China under Grant No.62176098. The computation is completed on the HPC Platform of Huazhong University of Science and Technology.
{
    \small
    \bibliographystyle{splncs04}
    \bibliography{main}
}
\appendix
\newpage

\section*{Appendix}                
\setcounter{subsection}{0}         
\renewcommand{\thesubsection}{\Alph{subsection}} 

\subsection{Additional Implementation Details}
\label{app:a}
\noindent\textbf{Architecture and Implementation Details}
We build DeCo on top of the pre-trained Stable Diffusion 3 (SD3). 
To decouple anomaly structures from product textures, we employ two distinct LoRA modules. 
The Product LoRA (P-LoRA) is configured with a rank of $r=64$. 
The Anomaly LoRA (A-LoRA) utilizes a full-rank configuration to accommodate the diverse and complex structures of anomalies. 
Both LoRAs are injected into the attention mechanisms of the MM-DiT blocks. 
The available target modules include the image-modality attention blocks (specifically, $W^0_\text{IQ},W^0_\text{IK},W^0_\text{IV},W^0_\text{IOut}$)
and their corresponding text-modality blocks.
Following the QK-Routing mechanism described in Sec.4.2 in the main text, A-LoRA and P-LoRA are selectively applied to different subsets of these modules for learning anomaly structure.

\noindent\textbf{Training Details.}
The training pipeline consists of two phases, both operating at an image resolution of $1024 \times 1024$. 
For the normal product, we train the P-LoRA for $1200$ steps with a batch size of $2$ and a constant learning rate of $1 \times 10^{-4}$. 
Separately, to acquire anomaly information, we train the A-LoRA for each reference product. This phase uses a batch size of $1$ and a learning rate of $5 \times 10^{-5}$. 
The total training steps for A-LoRA are dynamically set to $400 \times N$, where $N$ denotes the number of available real anomaly categories for the specific reference product.

\noindent\textbf{Inference and Sampling.}
During inference, images are generated at a resolution of $1024 \times 1024$ using the FlowMatchEulerDiscreteScheduler. The denoising process is completed in $25$ inference steps. We apply a Classifier-Free Guidance (CFG) scale of $5.0$.

\noindent\textbf{Computational Cost.}
We evaluate the computational cost of DeCo on a single NVIDIA A100 GPU (Tab.~\ref{tab:efficiency}). While AnomalyAny~\cite{sun2025unseen} and AnoStyler~\cite{so2026anostyler} are training-free, they suffer from slow \textit{on-the-fly} optimization during inference. In contrast, since UnzipLoRA~\cite{liu2025unziplora}, QR-LoRA~\cite{yang2025qr}, and DeCo all utilize dual-LoRA architectures, their training costs are comparable. 
For DeCo, training a P-LoRA requires 30 minutes, while each A-LoRA takes 10 minutes per anomaly type. 
During inference, the LoRA weights are statically merged, requiring no additional optimization. 
The minimal inference overhead of 4.5s is primarily attributed to the Product Compatibility Correction (PCC) module.

\begin{table}[h]
\centering
\caption{Computational Cost.}
\vspace{-5pt}
\label{tab:efficiency}
\tabcolsep=15pt
\renewcommand{\arraystretch}{1.1}
\resizebox{0.9\linewidth}{!}{%
\begin{tabular}{l|ccc}
\toprule\toprule
\textbf{Method} & \textbf{Train Time} (min/cls) & \textbf{Infer Time} (s/img) & \textbf{VRAM} (GB) \\
\midrule
AnomalyAny~\cite{sun2025unseen} & -- & 120.0 & 38 \\
AnoStyler~\cite{so2026anostyler}  & -- & 23.0  & 8 \\
QR-LoRA~\cite{yang2025qr}    & $30 + k \times 11$ & 4.0 & 23 \\
UnzipLoRA~\cite{liu2025unziplora}  & $30 + k \times 12$ & 4.0 & 23 \\
\midrule
\textbf{Ours (DeCo)} & $30 + k \times 10$ & 4.5 & 23 \\
\bottomrule\bottomrule
\end{tabular}%
}
\end{table}

\subsection{Additional Ablation Studies}
\label{app:b}
\begin{table}[t]
\centering
\renewcommand{\arraystretch}{1.3}
\tabcolsep=3pt
\begin{minipage}{0.48\linewidth}
    \centering
    \caption{\textbf{Ablation on A-LoRA Rank.} Comparison of a limited rank ($r=64$) and full-rank configurations for A-LoRA.}
    \label{tab:ablation_rank}
    \resizebox{\linewidth}{!}{%
    \begin{tabular}{l|cccccc}
    \toprule\toprule
    \multicolumn{1}{c|}{\multirow{2}{*}{\textbf{A-LoRA Config.}}} & \multicolumn{3}{c}{\textbf{Image-Level}} & \multicolumn{3}{c}{\textbf{Pixel-Level}} \\
    \cmidrule(lr){2-4} \cmidrule(lr){5-7}
     & \textbf{I-AUC} & \textbf{I-AP} & \textbf{I-F1} & \textbf{P-AUC} & \textbf{P-AP} & \textbf{P-F1} \\
    \midrule
    Rank 64 & 97.0 & 98.8 & 96.7 & 95.2 & 65.2 & 62.8 \\
    \textbf{Full Rank (Ours)} & \textbf{98.0} & \textbf{99.1} & \textbf{96.7} & \textbf{95.3} & \textbf{68.0} & \textbf{64.5} \\
    \bottomrule\bottomrule
    \end{tabular}%
    }
\end{minipage}
\hfill
\begin{minipage}{0.48\linewidth}
    \centering
    \caption{\textbf{Ablation on AP-Routing.} Analysis of different text routing strategies.}
    \label{tab:ablation_ap_routing}
    \resizebox{\linewidth}{!}{%
    \begin{tabular}{l|cccccc}
    \toprule\toprule
    \multicolumn{1}{c|}{\multirow{2}{*}{\textbf{Strategy}}} & \multicolumn{3}{c}{\textbf{Image-Level}} & \multicolumn{3}{c}{\textbf{Pixel-Level}} \\
    \cmidrule(lr){2-4} \cmidrule(lr){5-7}
     & \textbf{I-AUC} & \textbf{I-AP} & \textbf{I-F1} & \textbf{P-AUC} & \textbf{P-AP} & \textbf{P-F1} \\
    \midrule
    w/o Routing   & 95.1 & 97.9 & 96.3 & 94.9 & 66.4 & 62.8 \\
    w/o P-Routing & 96.7 & 98.5 & 96.3 & 95.3 & 67.4 & 63.9 \\
    w/o A-Routing & 96.9 & 98.0 & 96.3 & 95.0 & 66.9 & 63.0 \\
    \textbf{Ours} & \textbf{98.0} & \textbf{99.1} & \textbf{96.7} & \textbf{95.3} & \textbf{68.0} & \textbf{64.5} \\
    \bottomrule\bottomrule
    \end{tabular}%
    }
\end{minipage}
\vspace{-15pt}
\end{table}

\noindent\textbf{Impact of A-LoRA Rank.}
As shown in Tab.~\ref{tab:ablation_rank}, reducing the A-LoRA rank to $64$ degrades the pixel-level performance. 
This drop is mainly due to the large semantic gap between SD3's pre-training data and the industrial anomaly structure. Normal product textures are usually repetitive, making them easy for a low-rank LoRA to learn. 
However, the structure of the anomaly is complex. 
Since pre-trained SD3 lacks prior knowledge of these specific anomalies, a low-rank LoRA does not have enough capacity to reconstruct them.
Therefore, we adopt a full-rank A-LoRA to accurately acquire the anomaly structure.

\noindent\textbf{Effectiveness of AP-Routing.}
We conduct an ablation study on AP-Routing in Tab.~\ref{tab:ablation_ap_routing}.
Removing AP-Routing entangles the anomaly and product tokens, which degrades the generation quality and harms downstream performance. 
Specifically, removing A-Routing (w/o A-Routing) means the anomaly token $\text{T}_{\text{A}}$ is no longer mapped solely by A-LoRA, which fails to explicitly bind the anomaly structure to the abnormal token.
Conversely, during training, we use P-LoRA to align the image product features with the product token $\text{T}_{\text{P}}$.
Removing P-Routing (w/o P-Routing) means the product token $\text{T}_{\text{P}}$ is not mapped solely by P-LoRA.
Consequently, the image-modality Keys and Values mapped by P-LoRA cannot precisely represent the product texture, which negatively affects the decoupling of anomaly structure and product texture. 
Our full AP-Routing maps $\text{T}_{\text{A}}$ and $\text{T}_{\text{P}}$ separately, ensuring each is processed solely by its respective LoRA, thereby achieving the best performance.

\begin{table}[!t]
\centering
\renewcommand{\arraystretch}{1.3}
\tabcolsep=3pt

\begin{minipage}[t]{0.5\linewidth}
    \centering
    \caption{\textbf{Ablation on QK-Routing.} Analysis of different image token routing strategies.}
    \vspace{-5pt}
    \label{tab:ablation_aca}
    \resizebox{\linewidth}{!}{%
    \begin{tabular}{l|cccccc}
    \toprule\toprule
    \multicolumn{1}{c|}{\multirow{2}{*}{\textbf{Strategy}}} & \multicolumn{3}{c}{\textbf{Image-Level}} & \multicolumn{3}{c}{\textbf{Pixel-Level}} \\
    \cmidrule(lr){2-4} \cmidrule(lr){5-7}
     & \textbf{I-AUC} & \textbf{I-AP} & \textbf{I-F1} & \textbf{P-AUC} & \textbf{P-AP} & \textbf{P-F1} \\
    \midrule
    Symmetric & 94.8 & 97.6 & 95.8 & 93.5 & 63.1 & 61.6 \\
    Hybrid & 97.1 & 98.6 & 96.3 & 94.5 & 64.3 & 62.0 \\
    \textbf{Ours} & \textbf{98.0} & \textbf{99.1} & \textbf{96.7} & \textbf{95.3} &\textbf{ 68.0} & \textbf{64.5} \\
    \bottomrule\bottomrule
    \end{tabular}%
    }
\end{minipage}
\hfill
\begin{minipage}[t]{0.48\linewidth}
    \centering
    \caption{\textbf{Impact of $\lambda_\text{PI}$.} Analysis of different loss weights of $\mathcal{L}_{PI}$ (the default value is 0.1).}
    \vspace{-5pt}
    \label{tab:ablation_rvc}
    \resizebox{\linewidth}{!}{%
    \begin{tabular}{c|cccccc}
    \toprule\toprule
    \multicolumn{1}{c|}{\multirow{2}{*}{\textbf{Scale $\lambda$}}} & \multicolumn{3}{c}{\textbf{Image-Level}} & \multicolumn{3}{c}{\textbf{Pixel-Level}} \\
    \cmidrule(lr){2-4} \cmidrule(lr){5-7}
     & \textbf{I-AUC} & \textbf{I-AP} & \textbf{I-F1} & \textbf{P-AUC} & \textbf{P-AP} & \textbf{P-F1} \\
    \midrule
    0 (w/o) & 97.5 & 98.9 & 96.5 & \textbf{95.3} & 67.2 & 63.6 \\
    0.01 & 97.8 & 99.0 & 96.5 & 95.2 & 67.6 & 64.2 \\
    \textbf{0.1} & \textbf{98.0} & \textbf{99.1} & \textbf{96.7} & \textbf{95.3} &\textbf{ 68.0} & \textbf{64.5} \\
    1.0 & 98.0 & 99.0 & 96.6 & 95.0 & 67.4 & 63.9 \\
    \bottomrule\bottomrule
    \end{tabular}%
    }
\end{minipage}
\vspace{-10pt}
\end{table}

\noindent\textbf{Ablation on QK-Routing.}
To validate our QK-Routing mechanism, we evaluate three different strategies between A-LoRA and P-LoRA.
Symmetric (both LoRAs applied to all projections), 
Hybrid ($Q_{\text{I}}$ via A-LoRA; $K_{\text{I}}$ and $V_{\text{I}}$ use both), 
and Ours (the proposed QK-Routing). 
As shown in Tab.~\ref{tab:ablation_aca}, our design outperforms all alternatives. 
Strategies that allow A-LoRA to leak into the $V_{\text{I}}$ projection (Symmetric and Hybrid) suffer substantial drops in pixel-level precision. 
In these setups, A-LoRA directly alters the feature content in $V_{\text{I}}$, which interferes with the expression of the product texture. 
This negatively affects the decoupling of anomaly structure and product texture. 
In contrast, we strictly restrict A-LoRA to $Q_{\text{I}}$. 
This design limits the available textures to the normal product and ensures that A-LoRA solely extracts the structure of the anomaly, 
achieving the highest localization accuracy (68.0\% P-AP and 64.5\% P-F1).

\noindent\textbf{Effect of PI-Flow Weight $\lambda_{\text{PI}}$.}
Tab.~\ref{tab:ablation_rvc} shows the impact of the PI-Flow constraint. 
Without it ($\lambda_{\text{PI}} = 0$), the abnormal token incorrectly binds to the product texture, which negatively affects the pixel-level precision. 
Adding a small weight ($\lambda_{\text{PI}} = 0.01$) slightly improves the results, but the constraint remains too weak. Setting $\lambda_{\text{PI}} = 0.1$ effectively limits this incorrect binding, ensuring the token captures the anomaly structure and achieving the best P-AP (68.0\%).
However, a strong constraint ($\lambda_{\text{PI}} = 1.0$) harms performance because it overwhelms the primary DR-Flow objective. 
We thus set $\lambda_{\text{PI}} = 0.1$ as our default.

\begin{table}[!t]
\centering
\caption{\textbf{Impact of PCC Scalar $\omega$.} Analysis of different correction degrees.}
\vspace{-5pt}
\label{tab:ablation_pcc}
\tabcolsep=13pt
\renewcommand{\arraystretch}{0.9}
\resizebox{0.8\linewidth}{!}{%
\begin{tabular}{c|cccccc}
\toprule\toprule
\multicolumn{1}{c|}{\multirow{2}{*}{\textbf{Scalar $\omega$}}} & \multicolumn{3}{c}{\textbf{Image-Level}} & \multicolumn{3}{c}{\textbf{Pixel-Level}} \\
\cmidrule(lr){2-4} \cmidrule(lr){5-7}
 & \textbf{I-AUC} & \textbf{I-AP} & \textbf{I-F1} & \textbf{P-AUC} & \textbf{P-AP} & \textbf{P-F1} \\
\midrule
1 (w/o) & 97.5 & 99.0 & 96.4 & 95.1 & 67.6 & 63.2 \\
\textbf{2} & \textbf{98.0} & \textbf{99.1} & \textbf{96.7} & \textbf{95.3} & \textbf{68.0} & \textbf{64.5} \\
3 & 96.9 & 98.6 & 96.5 & 95.1 & 67.0 & 63.9 \\
\bottomrule\bottomrule
\end{tabular}%
}
\vspace{-15pt}
\end{table}

\noindent\textbf{Effect of PCC Scalar $\omega$.}
Tab.~\ref{tab:ablation_pcc} ablates the scalar hyperparameter $\omega$ in the Product Compatibility Correction (PCC). Setting $\omega = 1$ is equivalent to removing PCC entirely ($\hat{v}_{\text{inf}} = \hat{v}_{\text{DR}}$). 
Without correction, the injected anomaly structure often exhibits incompatibility with the specific product, making the anomaly barely visible. 
This limits the downstream detection performance, yielding a P-AP of 67.6\%. 
Setting $\omega = 2$ effectively pushes the prediction away from the product-only prediction $\hat{v}_p$. 
This successfully compensates for the incompatibility, ensures clear anomaly generation, and achieves the highest downstream metrics (98.0\% I-AUC and 68.0\% P-AP). 
However, an excessively large scalar (e.g., $\omega = 3$) pushes the prediction too far. 
This over-correction aggressively suppresses the normal product features, disrupting the recoupling process. 
Consequently, the downstream P-AP drops to 67.0\%. 
Therefore, we adopt $\omega = 2$ as the optimal default.

\noindent\textbf{Robustness to Prompt.}
To evaluate the sensitivity of DeCo to textual inputs, we test diverse prompt templates: (i) \textit{``a <product> with an <anomaly>''} and (ii) \textit{``a close-up inspection image of a <product> suffering from a severe <anomaly>''}. As shown in Tab.~\ref{tab:ablation_prompt}, DeCo maintains consistent performance across both image-level and pixel-level metrics regardless of the template complexity. This demonstrates that our method is robust to textual variations.

\begin{table}[!t]
\centering
\caption{Robustness to prompt variations on the MVTec AD dataset.}
\vspace{-5pt}
\label{tab:ablation_prompt}
\tabcolsep=20pt 
\renewcommand{\arraystretch}{1.1}
\resizebox{\linewidth}{!}{%
\begin{tabular}{l|ccc|ccc}
\toprule\toprule
\multirow{2}{*}{\textbf{\large Prompt}} & \multicolumn{3}{c|}{\textbf{Image-Level}} & \multicolumn{3}{c}{\textbf{Pixel-Level}} \\
\cmidrule(lr){2-4} \cmidrule(lr){5-7}
 & \textbf{I-AUC} & \textbf{I-AP} & \textbf{I-F1} & \textbf{P-AUC} & \textbf{P-AP} & \textbf{P-F1} \\
\midrule
(i)  & 98.2 & 98.6 & 96.7 & 95.3 & 68.1 & 64.5 \\
(ii) & 97.9 & 98.8 & 97.0 & 95.4 & 67.8 & 64.4 \\
\midrule
\textbf{Ours (Default)} & 98.0 & 99.1 & 96.7 & 95.3 & 68.0 & 64.5 \\
\bottomrule\bottomrule
\end{tabular}%
}
\end{table}

\subsection{Generalization to the Real-IAD Dataset}
\label{app:c}
We conduct additional experiments on the challenging Real-IAD~\cite{wang2024real}. 
Real-IAD is a large-scale, multi-view industrial anomaly detection dataset characterized by significant intra-class variations. 
Following the cross-dataset evaluation protocol used for the VisA, we recouple the anomaly structures acquired from the MVTec AD reference set with the target products in Real-IAD. As shown in Tab.~\ref{tab:realiad_results}, DeCo consistently outperforms existing state-of-the-art methods across both image-level and pixel-level metrics. Notably, our method achieves 20.6\% in P-AP and 31.6\% in P-F1, significantly surpassing the baselines. This verifies the robust generalization of our structural decoupling and recoupling strategy when transferred to unseen, highly challenging industrial environments.

\begin{table}[!t]
\centering
\caption{Generalization performance on the Real-IAD dataset.}
\vspace{-5pt}
\label{tab:realiad_results}
\tabcolsep=18pt
\renewcommand{\arraystretch}{1.1}
\resizebox{\linewidth}{!}{%
\begin{tabular}{l|ccc|cccc}
\toprule\toprule
\multirow{2}{*}{\textbf{\large Method}} & \multicolumn{3}{c|}{\textbf{Image-Level}} & \multicolumn{4}{c}{\textbf{Pixel-Level}} \\
\cmidrule(lr){2-4} \cmidrule(lr){5-8}
 & \textbf{I-AUC} & \textbf{I-AP} & \textbf{I-F1} & \textbf{P-AUC} & \textbf{P-AP} & \textbf{P-F1} & \textbf{PRO} \\
\midrule
DRAEM~\cite{zavrtanik2021draem}     & 60.1 & 39.3 & 48.3 & 89.2 & 10.3 & 20.1 & 71.3 \\
AnoStyler~\cite{so2026anostyler} & 60.2 & 40.1 & 49.2 & 90.5 & 13.4 & 24.2 & 75.1 \\
\midrule
\textbf{Ours (DeCo)} & \textbf{67.3} & \textbf{43.1} & \textbf{53.1} & \textbf{94.2} & \textbf{20.6} & \textbf{31.6} & \textbf{83.5} \\
\bottomrule\bottomrule
\end{tabular}%
\vspace{-20pt}
}
\end{table}

\subsection{Additional Qualitative Results}
\label{app:d}
\begin{figure}[h]
\centering
\includegraphics[width=0.94\textwidth]{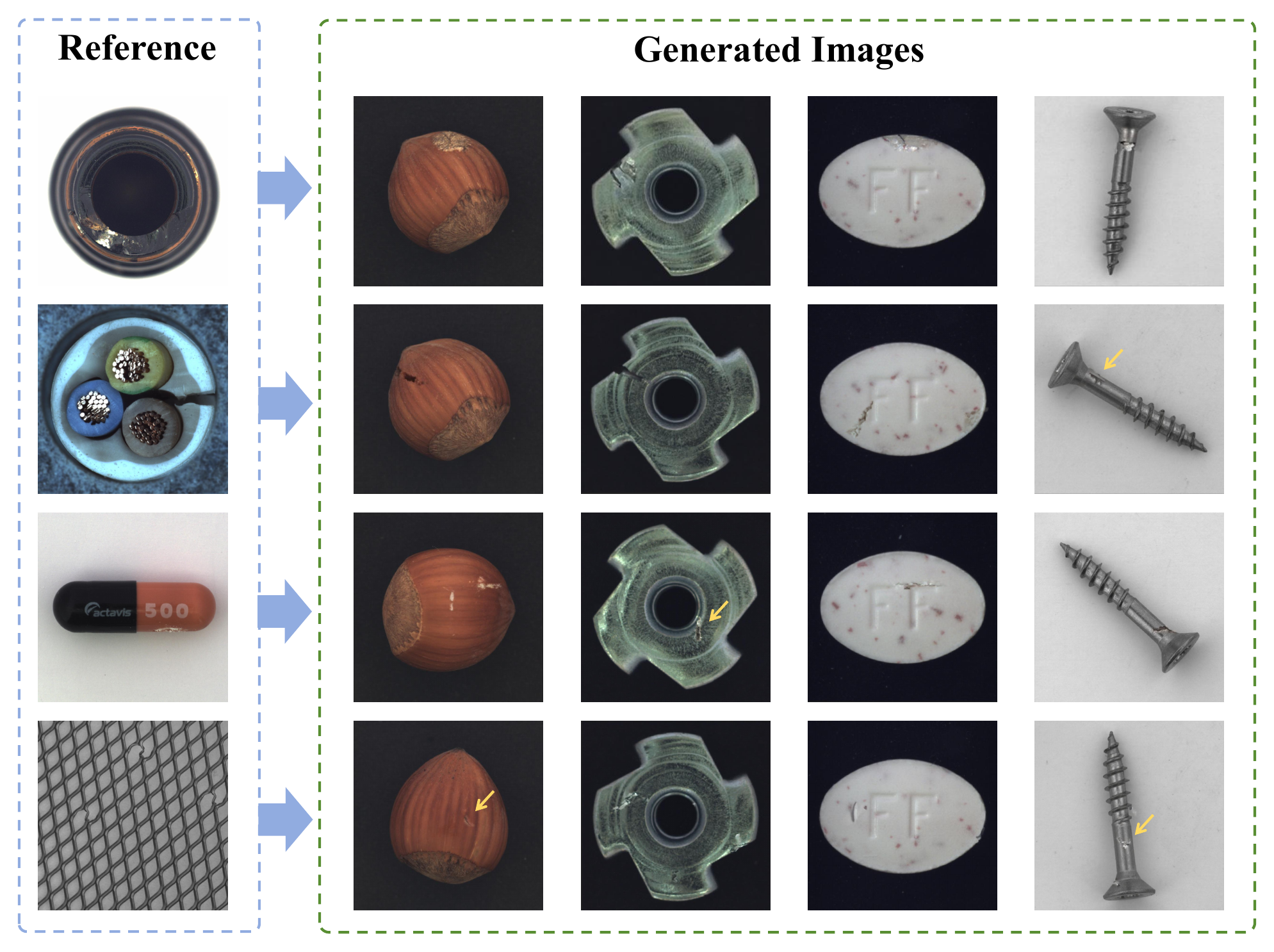}
\vspace{-10pt}
\caption{\textbf{Qualitative results of generated anomalies.} The leftmost column shows the reference anomalies, and the right columns show the generated images.}
\vspace{-20pt}
\label{fig:img1}
\end{figure}

\begin{figure}[h]
\vspace{-20pt}
\centering
\includegraphics[width=0.94\textwidth]{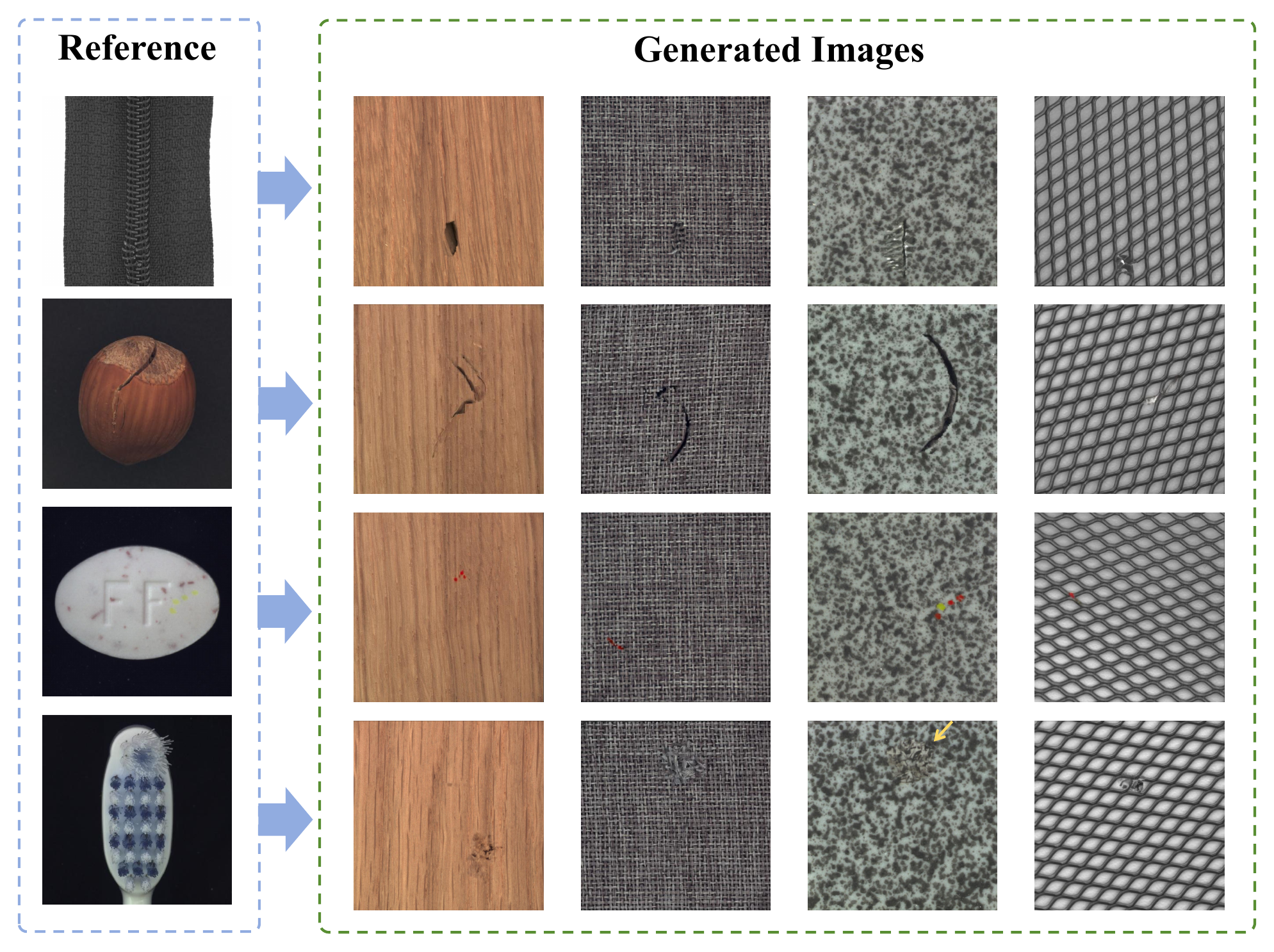}
\vspace{-10pt}
\caption{\textbf{More generated images.} The layout follows Fig.~\ref{fig:img1}.}
\vspace{-160pt}
\label{fig:img2}
\end{figure}

\begin{figure}[h]
\centering
\includegraphics[width=0.94\textwidth]{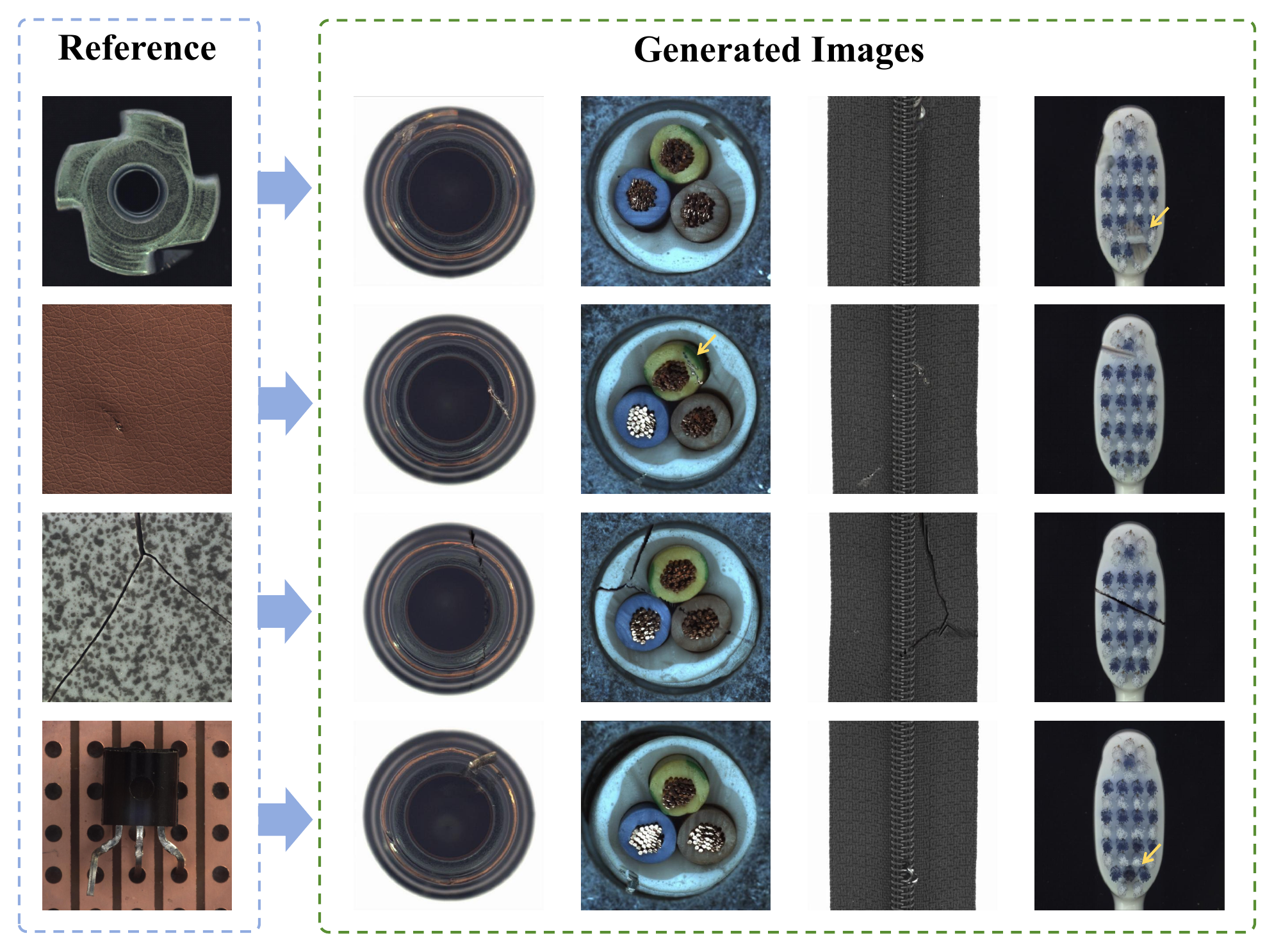}
\vspace{-10pt}
\caption{\textbf{More generated images.} The layout follows Fig.~\ref{fig:img1}.}
\vspace{-20pt}
\label{fig:img3}
\end{figure}

\begin{figure}[h]
\vspace{-20pt}
\centering
\includegraphics[width=0.94\textwidth]{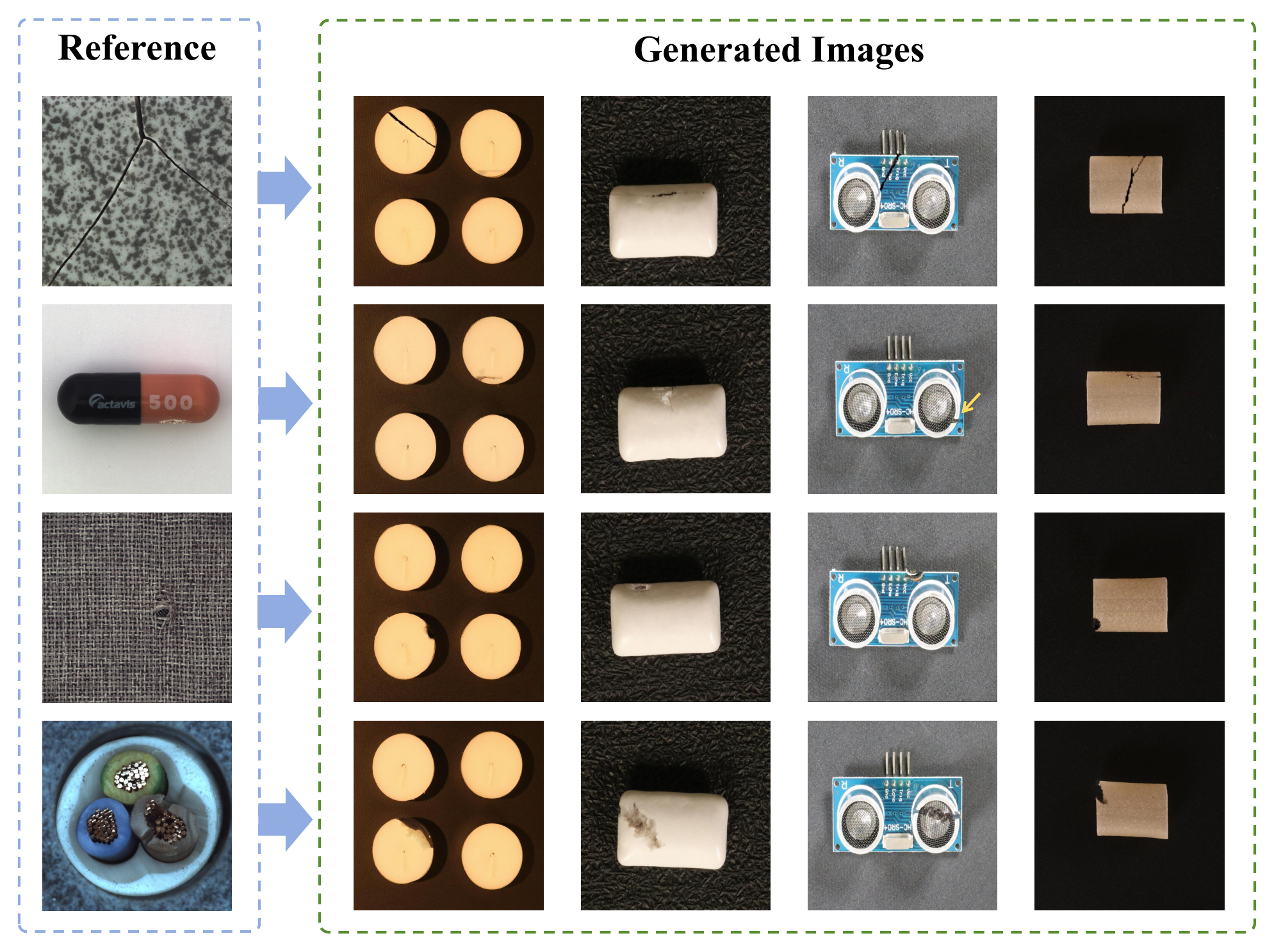}
\vspace{-10pt}
\caption{\textbf{More generated images.} The layout follows Fig.~\ref{fig:img1}.}
\vspace{-160pt}
\label{fig:img4}
\end{figure}

\begin{figure}[h]
\centering
\includegraphics[width=0.94\textwidth]{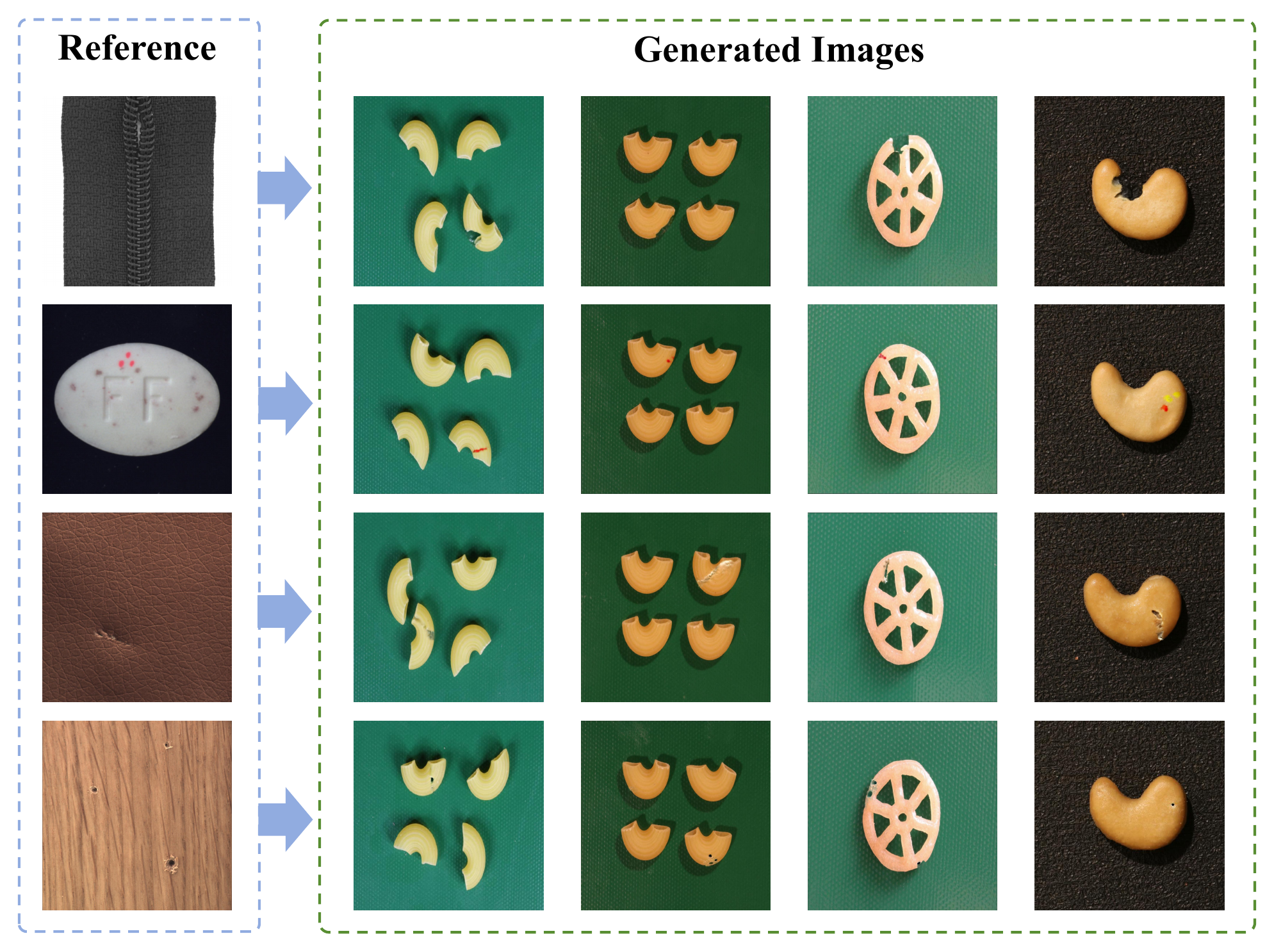}
\vspace{-10pt}
\caption{\textbf{More generated images.} The layout follows Fig.~\ref{fig:img1}.}
\vspace{-20pt}
\label{fig:img5}
\end{figure}

We provide more qualitative results, as shown in Fig.~\ref{fig:img1} to Fig.~\ref{fig:img5}. 
In each figure, the leftmost column displays the reference anomaly images. 
The corresponding images in the right columns show the generated images. 
As observed, our method successfully extracts the pure anomaly structure from reference anomalies of non-target products, and seamlessly recouples it with novel target product textures. 
These results visually confirm the strict decoupling of anomaly structure and product texture, as well as their successful recoupling.

\newpage

\newpage

\newpage

\newpage

\end{document}